# An improved wind power prediction via a novel wind ramp identification algorithm

**Authors:** Yifan Xu

**Abstract**: Conventional wind power prediction methods often struggle to provide accurate and reliable predictions in the presence of sudden changes in wind speed and power output. To address this challenge, this study proposes an integrated algorithm that combines a wind speed mutation identification algorithm, an optimized similar period matching algorithm and a wind power prediction algorithm. By exploiting the convergence properties of meteorological events, the method significantly improves the accuracy of wind power prediction under sudden meteorological changes. Firstly, a novel adaptive model based on variational mode decomposition, the VMD-IC model, is developed for identifying and labelling key turning points in the historical wind power data, representing abrupt meteorological environments. At the same time, this paper proposes Ramp Factor (RF) indicators and wind speed similarity coefficient to optimize the definition algorithm of the current wind power ramp event (WPRE). After innovating the definition of climbing and denoising algorithm, this paper uses the Informer deep learning algorithm to output the first two models as well as multimodal data such as NWP numerical weather forecasts to achieve accurate wind forecasts. The experimental results of the ablation study confirm the effectiveness and reliability of the proposed wind slope identification method. Compared with existing methods, the proposed model exhibits excellent performance and provides valuable guidance for the safe and cost-effective operation of power systems.

**Keywords**: VMD Composition; Wind power ramp event; Similar period matching; Deep Learning

## 1 Introduction

### 1.1 Motivation

In recent years, the "dual-carbon strategy" has emphasized green and sustainable development as a cornerstone of its implementation, with wind power serving as one of the core components of new energy. Globally, wind power has achieved significant progress. According to the *2024 Global Wind Energy Report* released by the Global Wind Energy Council (GWEC), 2024 marked the most successful year in the history of the wind power industry, with installed capacity increasing by 50% year-on-year. The record-breaking 117 GW of grid-connected wind capacity added in a single year highlights the industry's remarkable resilience and adaptability, demonstrating significant progress in the global fight against climate change. This growth included 106 GW of newly installed onshore wind power, representing a 54% year-on-year increase, and 10.8 GW of offshore wind power, reflecting a 24% year-on-year rise. By the end of 2023, the cumulative global wind power installed capacity had reached an impressive 1 TW, with onshore wind power accounting for 93% and offshore wind power comprising the remaining 7%[1]. Wind power, however, inherits the inherent intermittency, fluctuations, and stochastic nature of wind energy, necessitating a reliable wind power prediction method to support its large-scale integration into power grids.

Currently, wind power forecasting research still faces a key challenge: the low reliability of power forecasts during Wind Power Ramping Events[2]. Wind Power Ramping Events (WPRE) represent an extreme manifestation of wind power's stochasticity and volatility[3], and is one of the hot issues of current researchers in the field of wind power, which occurs in a short period of time and leads to large fluctuations in wind power generation[4]. The sudden increase or decrease of wind power in a short time is often caused by the sudden change of wind speed in convective weather, transient weather, and other sudden weather changes. Studies have shown that wind slope events are one of the important factors causing grid security accidents and are also a major factor affecting the accuracy of wind power prediction[5]. For example, California's wind power dropped from 4,000 MW to zero overnight in 2014[6], Germany's wind output surged to 40 GW during a 2017 storm [7], and a UK storm in 2019 caused a spike to 25 GW[8]. Such events are not limited to these regions; similar situations have been observed globally. Addressing the inaccuracies in wind power prediction caused by ramping events is critical to mitigating the current energy crisis and ensuring grid reliability[9].

**1.2 Literature review**

The time scale of short-term wind power prediction typically ranges from 1 to 3 days. To enhance prediction accuracy, precise wind speed forecasting is essential. Existing methods for wind speed prediction can be broadly classified into two categories: data-driven approaches and physical models. Among these, data-driven approaches have gained significant attention from researchers, leveraging intrinsic correlations within meteorological data from diverse sources to construct mathematical-statistical model[10]-[11], machine learning models [12]-[13], and hybrid models[14]-[15]. However, due to the constraints that the prediction step size provided by statistical schemes is difficult to meet the needs of wind farms[16], and that it is difficult for wind power plants to meet the data requirements of the statistical method, the physical scheme, i.e., numerical weather prediction, is usually used as the source of meteorological data for the wind power prediction model in wind power farms .

Consequently, physical methods, particularly numerical weather prediction (NWP), are widely used as the meteorological data foundation for wind power forecasting. NWP employs physical equations, such as those derived from hydrodynamics and thermodynamics, to simulate atmospheric dynamics within defined temporal and spatial domains. Prominent physical models for wind speed prediction include the High-Resolution Limited Area Model (HIRLAM)[17], the Mesoscale Model5 (MM5)[18] and the Weather Research and Forecasting (WRF) model [19]. Among them, the WRF model can provide ultra-high resolution and its Large Eddy Simulations (LES) scheme can effectively simulate the atmospheric motions of wind farms, which is favored by wind farm operators [20]. However, physical models are prone to significant errors stemming from inadequate parameterization, low resolution, and inaccurate topographic representation[21]. Due to the errors in wind speed prediction, it is difficult to meet the needs of most wind farms [22]-[23]. Consequently, standalone physical models often require supplementation by other methods to achieve the high levels of accuracy demanded in practical engineering contexts.

In contrast, data-driven models integrate meteorological data with wind farm power output to predict wind power. This category includes time series models, machine learning algorithms, and hybrid approaches. Traditional time series models, such as autoregressive moving average (ARMA)[24], Kalman filter[25] , Markov chain [26] and autoregressive integrated moving average model (ARIMA) [27], can effectively satisfy the demand of wind power prediction for wind farms under normal smooth conditions, but it is difficult to deal with the complex meteorological conditions of wind power prediction. Their linear nature limits their effectiveness in capturing the complexity of wind dynamics under turbulent conditions. With developments in artificial intelligence, machine learning models have emerged as powerful tools for wind power prediction due to their nonlinear fitting capabilities. Similarly, deep learning models are gaining prominence for their ability to process multidimensional data and deliver superior prediction accuracy[28],[29],[30]. Obviously, hybrid models and deep learning algorithms have been widely used in wind speed and power prediction modeling and have produced better results.

In summary, the research on wind power prediction methods has made great development, and the related products have shown excellent market prospects. However, due to the defects of low completeness of initial meteorological field and poor quality of meteorological observation data[31]  most of the errors in wind power prediction come from the wind speed prediction stage [32].

Currently, the phenomenon of sudden increase and decrease of wind speed in a short period of time is one of the hotspots for researchers, and this kind of problem can be attributed to the stochastic and fluctuating nature of wind power, and the traditional wind power prediction can hardly satisfy the demand for stable operation of the power grid during the sudden change of wind speed represented by the drastic increase or decrease of wind speed, and thus worldwide researchers have carried out the related research of wind power prediction for the sudden change of wind speed[33].

Currently, the sudden increase or decrease of wind speed, characterized by wind speed ramp events, has become a significant focus for researchers due to its inherent randomness and fluctuation. These events pose substantial challenges for traditional wind power prediction methods, which often struggle to meet the operational stability requirements of power grids during such abrupt changes. As a result, researchers worldwide have conducted extensive studies on wind power prediction under wind speed ramp conditions. Jie Li et al. developed a hybrid prediction model combining Self-Attention Mechanism (SAM) and Wavelet Transform (WT) with a one-dimensional convolutional neural network (1D CNN) and Long Short-Term Memory (LSTM) network to predict Wind Power Ramp Events (WPRE)[34]. Similarly, Cui et al. proposed an improved hybrid model leveraging LSTM by incorporating considerations of wind power ramp events to enhance predictive performance[7]. Jiang et al. introduced a day-ahead ramp event prediction method based on ramp event definition and profile analysis[35]. They proposed the Typical Event Clustering Identification (TECI) algorithm, which utilizes EB-K clustering and similarity metrics to analyze historical data and identify high-probability ramp events. Harsh S. presented a hybrid method for predicting ramp events in onshore, offshore, and hilly regions, integrating Discrete Wavelet Transform (DWT) with various learning algorithms such as Twin Support Vector Regression (TSVR), Random Forest Regression (RFR), and Convolutional Neural Networks (CNN) [36]. Although these methods demonstrate excellent performance in wind power prediction, most rely on fitting models solely based on wind power curves without fully utilizing the implicit information in wind speed data to improve prediction accuracy.

Based on the above, there remain areas for improvement in existing wind speed mutation power prediction methods:(1) Most of the current wind power prediction methods fit the wind power curve itself to a machine learning model or a deep learning model, which lacks the emerging ideas for prediction.(2) Current power prediction methods do not tap into the features present in the wind speed series itself, such as the correlation of the wind speed series. (3) Sudden wind speed increases or decreases due to sudden wind speed events reduce the accuracy of wind power predictions.

## 1.3 Contribution

In this study, a novel wind speed mutation event identification algorithm based on a similar period matching strategy is proposed. A new ramp identification method is developed, integrating the "similar day" concept from photovoltaic power prediction into wind power prediction. This approach extracts convergence features from historical meteorological data, while the Informer deep neural network, leveraging the Attention mechanism, is employed for multimodal wind power prediction. The novelty and contribution of this study are as follows:

(1) A dynamic adaptive method for identifying wind speed mutation events is introduced. This approach mitigates issues such as modal aliasing in decomposition results while accurately capturing the timing of wind speed mutation events. The method provides critical support for locating wind speed mutation events and enhances the precision of subsequent prediction models.

(2) The study incorporates the "similar day" concept from photovoltaic power prediction into wind power forecasting for wind speed mutation events. A novel similar period matching algorithm based on wind speed correlation is proposed. By referencing the target prediction period, this method identifies convergent meteorological data from historical records, enabling accurate predictions of wind speed and wind power following wind speed mutation events.

(3) The Informer deep learning model is employed to predict wind speed mutation events, addressing the computational complexity of the traditional Transformer model. This approach considers both local and global time series timestamps and integrates newly proposed multimodal data, such as the ramp factor and wind speed correlation. The result is an improved ability to predict wind speed and wind power during mutation events.

## 1.4 Paper organization

The remainder of this paper is as follows: chapter 2 introduces the improved VMD algorithm as well as the dynamic adaptive selection model, chapter 3 introduces the similar period matching method with the addition of the new climb definition and wind speed correlation definition, chapter 4 introduces the Informer deep learning algorithm, and chapter 5 gives the specific experimental results with a comparative analysis of the experiments. Finally, chapter 6 provides some conclusions and outlook for future research.

## 2 Data

### 2.1 Wind farm data

The research object of this paper is a wind farm in Baijiashe, Jiangxi Province as shown in Figure 1, its latitude and longitude coordinates (27°07'N,116°99'E), the region is located in the subtropical monsoon climate zone, the same period of water and heat in the summer and fall season is subject to tropical cyclones and other extreme weather effects. The wind farm is equipped with 20 wind turbines of type H151. This type of fan in the wind speed of 3.5m/s into the operation mode can be external power output, in the wind speed of 10.5m/s when the fan will run at full power, when the wind speed of 25m/s when the fan will be put on the paddle shutdown. The output power at full power is 5MW, the radius of the paddle is 76m, and the swept area of the paddle is 18146 m$^2$. The total installed capacity of the wind farm is 100 MW, and the hub height of the turbines is 90 m. In addition, meteorological information such as wind speed and direction at the hub height of the turbines in the wind farm is measured by an on-site anemometer tower, and the power data of each turbine is collected by a SCADA system installed at the tail end of the turbines. The sampling interval of all the data in this study is 15 min. Considering the long-term maintenance and testing of different wind turbines in this wind farm, the power prediction part is centered on a single long-term on-line operating wind turbine in the target wind farm.

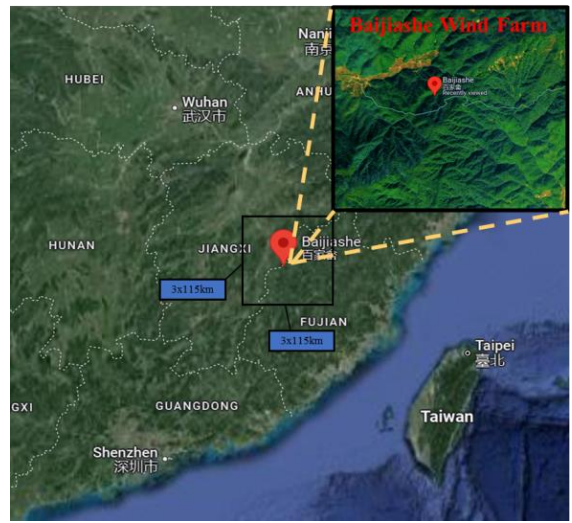

Figure 1. Geographic location of the wind farm

In this study, wind speed and power data of wind turbines in a wind farm in Baijiashe, Jiangxi Province were used, which mainly consisted of two parts, the past data $w^p$ and the historical data

$w^h$ in which the past wind speed data were selected as experimental data for the wind speed decomposition data of the wind farm from November 6, 2023 to June 20, 2024, and the historical wind speed data used all wind speed data of the Baijiashe wind farm in the historical database. The historical wind speed data used all the wind speed data in the historical database of Baijiashe wind farm in the previous 2 years from the past wind speed data.

**2.2 NWP data**

NWP stands for Numerical Weather Prediction, which is the modeling of weather and related variables such as temperature, humidity, wind speed, etc. over time through numerical models[37]. Numerical weather prediction through numerical modeling provides the evolution of weather variables over time and space. Such models can help predict weather conditions for the next few hours, days or even weeks, as well as predict the development and evolutionary trends of large-scale weather systems. The numerical weather prediction (NWP) data used in this paper are from the China Meteorological Administration (CMA) and are issued twice a day with 15-minute temporal resolution, 3.0-kilometer spatial resolution, and an expected length of 144 hours. The NWP elements include shortwave radiation value, radiation amount, air temperature at 2 m, relative humidity at 2 m, pressure, temperature at each height, and wind direction data and wind speed data at each height. Input features play a very important role in modeling, and the correlation matrix (Fig. 2) is often used as a statistical indicator to analyze the closeness of correlation between variables.

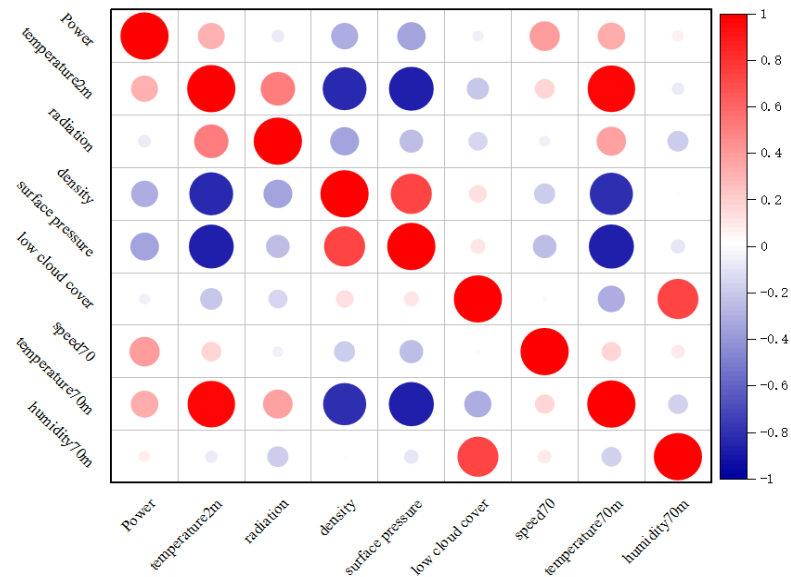

Figure 2. NWP correlation matrix

Among the predicted variables, the wind speed data at 70 m demonstrates the strongest positive correlation with historical wind power data, with a correlation coefficient of 0.40, followed by the temperature at 70 m with a correlation coefficient of 0.33. Conversely, surface pressure and density exhibit significant negative correlations with wind power data, with coefficients of -0.35 and -0.32, respectively. In this study, the two variables with the highest correlation were selected as model inputs. To address the varying scales of these input variables, they were standardized using the min-max normalization method prior to modeling.

## 2.3 Wind power ramp data

Wind power is closely correlated with wind speed, and most contemporary wind power prediction methods account for this relationship. The wind speed-power correlation curve, depicted in Fig. 3 under ideal operating conditions, represents an essential characteristic of the operational performance of wind turbines (WTGs).The power curve, often referred to as the performance curve, illustrates the relationship between wind speed (WS) and active power output. It is defined as a characteristic curve constructed from a series of specified data pairs (WS, Power), where wind speed serves as the horizontal axis and active power as the vertical axis. Under standard atmospheric density conditions, this relationship is referred to as the standard power curve of the WTG, serving as a critical benchmark for assessing turbine performance [38].

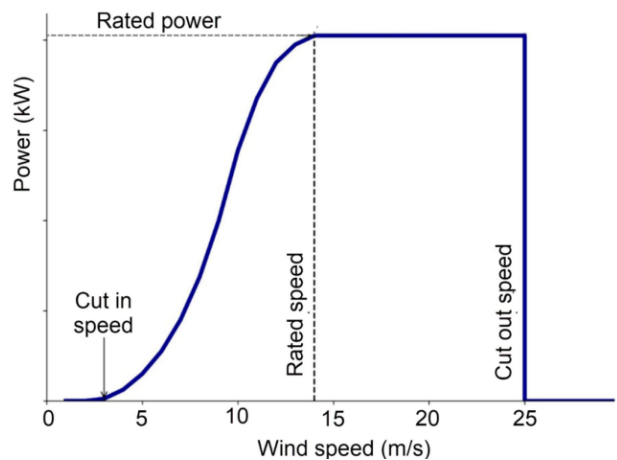

Figure 3. Wind Speed-Power Correlation Curve

The standard power curve does not account for the inherent volatility, randomness, and intermittency of wind. These characteristics of wind often result in rapid increases or decreases in wind speed over short periods, leading to corresponding abrupt changes in the power output of wind farms. Such fluctuations can significantly disrupt the stable operation of the power grid, potentially causing localized frequency reductions or even voltage collapse, thereby posing serious risks to grid reliability. Here we use part of the data of Jiangxi Baijiashe in 2024 to observe the regular operation status of local wind turbines and the abnormal status of the occurrence of ramp events. The details are shown in Fig.4. Fig.4 (a) shows the wind speed series of 150 15-minute data points at two different periods, Fig.4 (b) shows the comparison of the violin plots of the two, and Fig.4 (c) shows the comparison of the wind power curves corresponding to the two wind speed series. From the three figures, the ramp event in wind power prediction is prone to cause active power imbalance in the system, destroying the frequency stability and seriously threatening the safe, stable, and economic operation of the power grid.

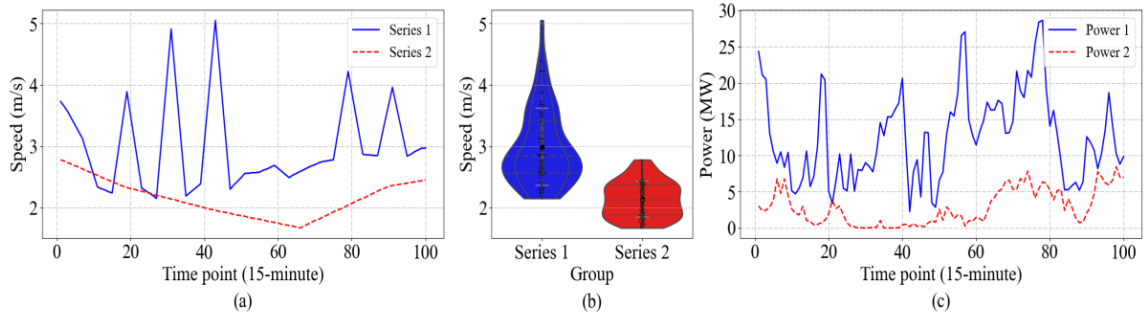

Figure 4. Wind speed and power conditions

# 3 Methods

## 3.1 Novel VMD-IC Algorithm

### 3.1.1 VMD

Considering that this study adopts a data-driven approach to wind power prediction, the accurate prediction of wind power during wind speed mutation events requires substantial historical meteorological data support. Wind speed mutation events are characterized by sharp increases or decreases in wind speed within a specific period, making the identification of these events a task focused on filtering out minor fluctuations while preserving the overarching trend of sharp variations. Currently, wind power researchers frequently employ the revolving door algorithm for such tasks. However, a key limitation of this method is the need for manual gate-width adjustment based on the intrinsic characteristics of the dataset, rendering the algorithm less suitable for handling large-scale datasets automatically and efficiently. This inefficiency introduces unnecessary computational and operational costs[39]. In this section, a VMD-based dynamic adaptive wind speed mutation event recognition method (later referred to as VMD-IC) is proposed, which consists of a VMD modal decomposition model, a pole adaptive selection model and a split-signal dynamic screening model, and the specific structure is shown in Figure 5.

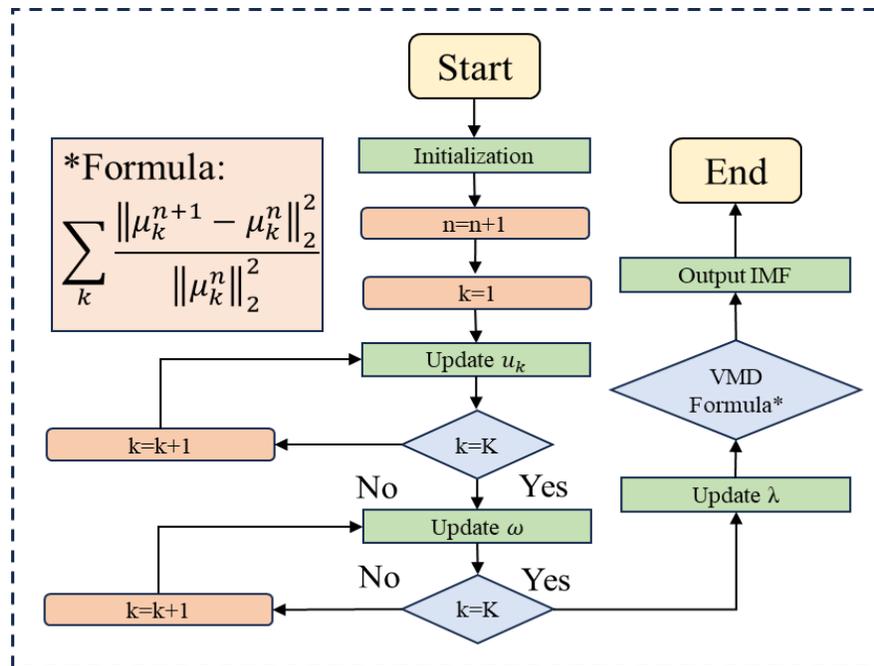

Figure 5. VMD modeling process

The variational mode decomposition (VMD) algorithm, a non-recursive adaptive signal processing tool, is specifically designed to decompose the original signal into a series of intrinsic mode components (IMCs), each characterized by a limited bandwidth and a specific center frequency. The VMD model iteratively optimizes the variational model to extract sub-signals that represent the optimal solution[40]. One of the core advantages of the VMD model is its ability to decompose the target signal into a predetermined number of sub-signals, denoted as K, which is specified by the operator. This adaptability ensures that the model effectively handles signals under varying and complex working conditions, demonstrating robust generalization capabilities[41].

Unlike the empirical mode decomposition (EMD) model, the variational mode decomposition (VMD) model allows the operator to manually specify the number of decomposed signals, effectively avoiding issues such as modal aliasing in the decomposition results. This feature ensures that the VMD model delivers superior and more stable performance in addressing non-smooth and non-linear complexities encountered in engineering applications[39]. Consequently, the VMD model is chosen as the foundational approach for identifying sudden meteorological events in this study.

### 3.1.2 Pole Adaptive Selection Model

The VMD modal decomposition model outputs the decomposition signal with small fluctuations as shown in Fig. 6. In the figure, the green solid circle is the very small value point, the red solid circle is the very large value point, the blue solid line is a section of the example wind speed sequence, and the red shaded area is the region of fluctuating pseudo-polar points under study, which is located from the 6110th time point to the 6150th time point, and, as shown by the partial zoom in the upper right corner, there are three close extreme value points and three very small value points within a short period of time, and it is found that the pseudo-polar point phenomenon described above occurs. As shown in the zoomed-in figure on the right, there are three close maxima and three minima within a short period of time, resulting in the pseudo-polar phenomenon described above. Meanwhile, the observation of the overall wind speed sequence reveals that the pseudo-polar phenomenon is not an accidental phenomenon with a low probability, and pseudo-poles are also present in the vicinity of the time points of 6200 and 6450, etc. The red shaded area is the studied fluctuation pseudo-polar area.

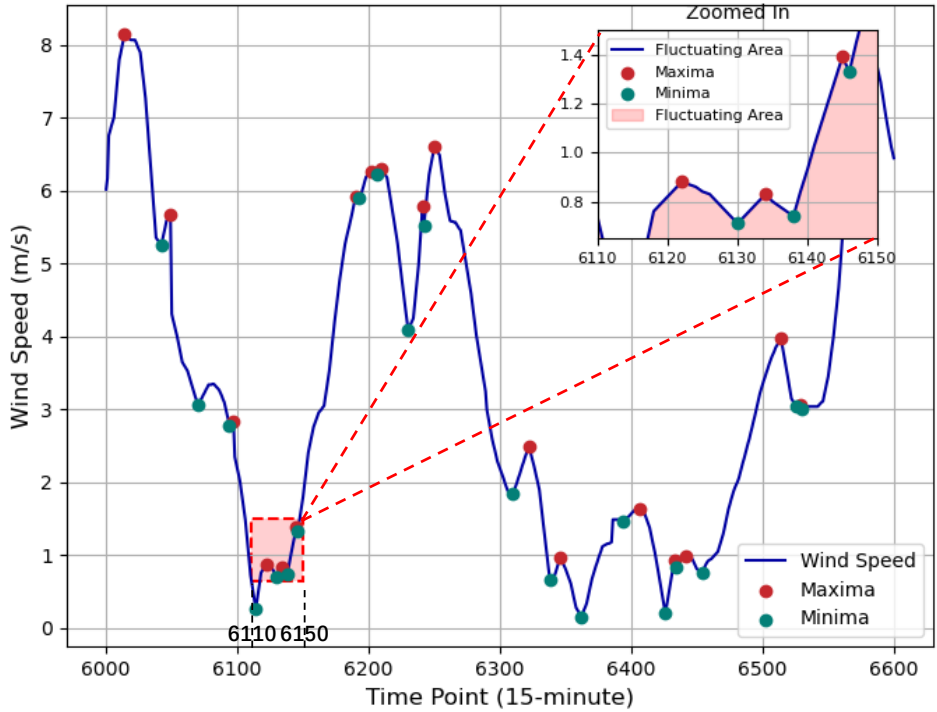

Figure 6. Pseudo-polar phenomenon of split signals.

The VMD modal decomposition produces a significant number of extreme points that align with the definition of poles. However, many of these exhibit minimal Euclidean distances between extremely large and small values, leading to overlapping neighboring extrema. This overlap significantly complicates the implementation of sub-signal selection methods based on the number of extreme points. To address this issue, this study proposes a pole adaptive selection model designed to resolve the problem of overlapping extrema within sub-signals. The implementation process is outlined as follows:

Step 1: Extract the maximum value $Z_{max}$, the minimum value $Z_{min}$ and the set of all extreme points $P = \{p_1, p_2, \cdots, p_n\}$ from the wind speed decomposition data, where $p_n$ denotes the nth extreme point;

Step 2: According to the adaptive coefficients and calculate the dynamic window width, the specific formula is:

$$width_{door} = \ell * |Z_{max} - Z_{min}| \tag{1}$$

In Equation (1), $width_{door}$ denotes the dynamic window width; $\ell$ denotes the adaptive coefficient, and the value of l is generally (0.03,0.08);

Step 3: According to the Euclidean distance between the near-neighbor poles, keep the poles that satisfy $d_i = |p_i - p_{i+1}| > width_{door}$, where $d_i$ denotes the Euclidean distance between two neighboring poles and $i \in (0, n)$; get the set of poles after selection $P_{renew} = \{p_1^{re}, p_2^{re}, \cdots, p_i^{re}, \cdots, p_m^{re}\}$, where $p_i^{re}$ denotes the ith extreme point after selection and $i \in [1, m]$. The selected pole set $P_{renew}$ generated above will be applied as the reconstructed pole set of wind speed decomposition data for the identification of wind speed mutation events, and its specific process is shown in Figure 7.

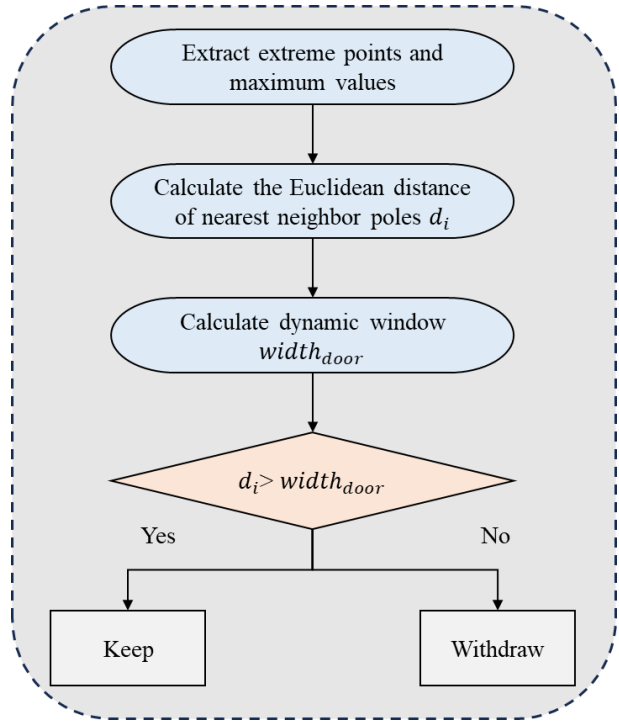

Figure 7. Pole selection model

### 3.1.3 Distributor signal screening model

This section introduces the sub-signal dynamic screening model, with the specific process illustrated in Fig. 8. The model calculates the pole rate $\eta$ based on the selected set of pole points derived from the pole adaptive selection model. By leveraging the number of selected pole sets, the dynamic screening model enables effective sub-signal data selection from the VMD modal decomposition results, facilitating accurate identification of wind speed mutation events. The calculation method is detailed as follows:

Step 1: Calculate the pole rate of wind speed data with the specific formula:

$$\eta_k = \frac{n_{renew}^k}{N_{original}} \tag{2}$$

In Equation (2), $\eta_k$ denotes the pole rate of the kth wind speed data, $N_{original}$ denotes the total number of reconstructed pole points of the preprocessed wind speed data, and $n_{renew}^k$ denotes the total number of reconstructed pole points of the kth wind speed data;

Step 2: filtering the wind speed data according to the pole rate of each wind speed data, retaining the wind speed data that satisfies $\eta_\kappa \leq \tau$, and obtaining the retained wind speed data set $imf_{reserved}$, where $\tau$ is a pole rate threshold set to 1;

Step 3: The retained wind speed data set $imf_{reserved}$ is superimposed to realize wind speed data reconstruction, and the reconstructed wind speed data is noted as $IMF_{renewd}$, with the specific formula:

$$IMF_{renewed} = sum(imf_{reserved}) \tag{3}$$

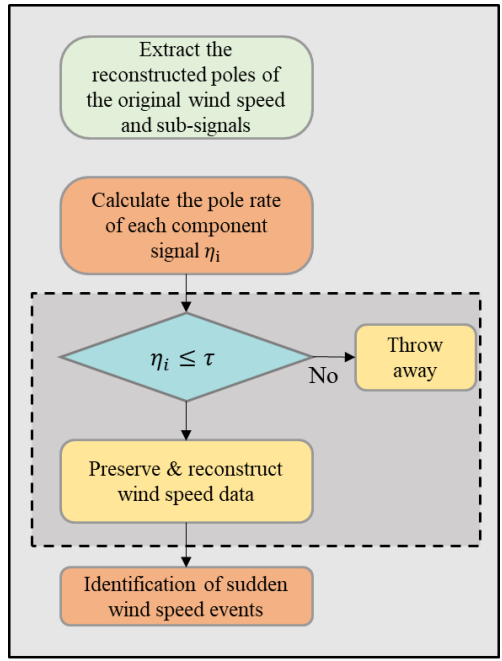

Figure 8. Sub-signal dynamic screening model

### 3.2 Informer Deep Learning Algorithm

Zhou et al. proposed a Transformer-based variant of the deep learning algorithm Informer in 2021, which streamlines the model structure and thus reduces the arithmetic demand while ensuring that the excellent performance of the model's feature extraction is not compromised[42].

Informer consists of the following three components, whose specific flow is shown in Figure 9. The functions and structures of each part are described below in terms of the three parts: input expression, encoder, and decoder.

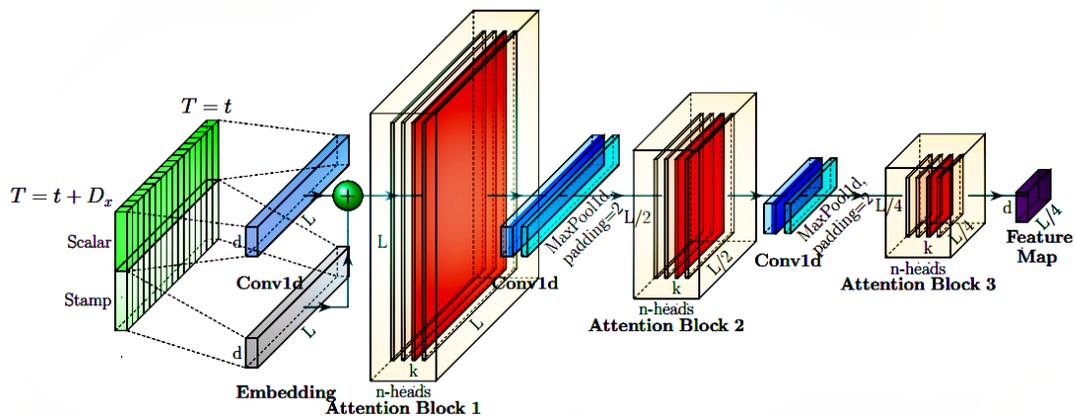

Figure 9. Informer Model Structure

### 3.2.1 Embedding

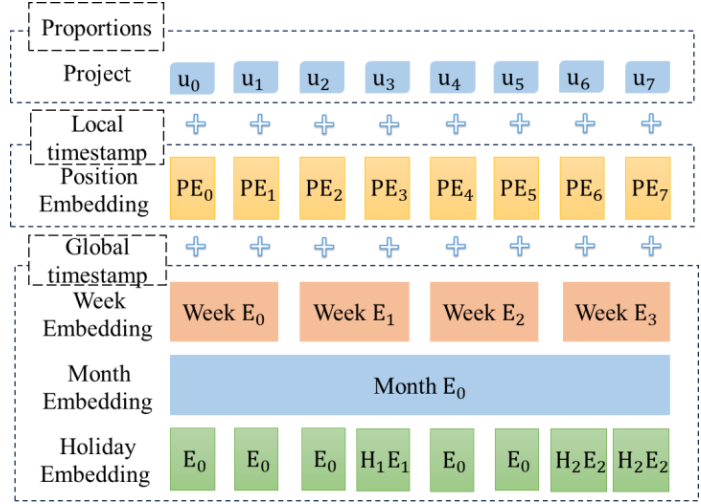

Figure 10. Input representation module structure

The Informer model's input representation module consists of three superimposed parts: scalar projection, local time stamps and global time stamps, as shown in Figure 10. For a scalar value of $x_i^t$ at time t in a time series $x_i$, the scalar projection is the projection of $x_i^t$ into a $d_{model}$-dimensional vector, denoted as $u_i^t$, using a one-dimensional convolutional filter with d_model convolutional kernel of size 3 and step size 1, where $d_{model}$ is the input and represents the dimensionality of the model. Time sequence In The local timestamp of the moment is encoded as the fixed position of the moment in the sequence:

$$PE(pos,j)^t = \begin{cases} \sin\left(\frac{pos}{2L_x}\right)^{\frac{j}{d_{model}}}, & if\ j\ is\ even \\ \cos\left(\frac{pos}{2L_x}\right)^{\frac{j}{d_{model}}}, & otherwise \end{cases} \quad (4)$$

where pos is an integer representing a fixed position in the sequence at moment t, $j = 1,2,\cdots,d_{model}$, and $L_x$ is the length of the input sequence, which takes the value of the encoder's and decoder's input sequences when representing their input sequences, respectively. The global timestamp is then the sum of four types of time: hour, day, week, and month, each of which $SE(pos)_p$ is computed by the nn. Embedding package in pytorch. Finally, the input sequence In The input of the time represents the result:

$$x_{feec[i]}^t = \alpha u_i + PE(L_x \times (t-1) + i) + \sum_{p=1}^{4} SE(L_x \times (t-1) + i)_p \qquad (5)$$
$$i = 1, 2, \cdots, L_x$$

In Equation (5), $\alpha \in [0,1]$ is the scalar projection coefficient.

### 3.2.2 Encoder

The encoder takes the fixed-length time series data before the prediction moment as the input of the Informer model, and applies two types of modules, namely, multi-head self-attention mechanism and self-attention distillation, to realize the mining and learning of large time-distance dependency features embedded in long sequences.

$$KL(a\|b) = \ln \sum_{j=1}^{N_K} e^{\frac{q_i k_j^T}{\sqrt{\sigma}}} - \frac{1}{L_K} \sum_{j=1}^{N_K} \frac{q_i k_j^T}{\sqrt{\sigma}} - \ln N_K \qquad (6)$$

In Equation (6), $q_i$ is the ith , $k_j$ is the jth key value, $N_k$ is the number of Keys, $p(k_i|q_i) = \frac{k(q_i,k_j)}{\sum_l k(q_i,k_l)}$ stands for the distribution of $q_i$ self-attentive distribution and $k(q_i, k_j) = e^{\frac{q_i k_j^T}{\sqrt{\sigma}}}$, and $b = 1/N_k$ is a uniform distribution. Removing the constant term in Eq. (28), the dominant query selection criterion in the Informer model is:

$$M(q_i, K) = \ln \sum_{j=1}^{N_K} e^{\frac{q_i k_j^T}{\sqrt{\sigma}}} - \frac{1}{L_K} \sum_{j=1}^{N_K} \frac{q_i k_j^T}{\sqrt{\sigma}} \qquad (7)$$

The first term in Equation (7) is the computation of the Log-Sum-Exp form of $q^i$ with all the key values under the parameter d, and the second term is their arithmetic mean. The higher value of $M(q_i, K)$ represents the higher probability that $q^i$ contains the dominant dot product pair. However, the computational complexity of Eq. (7) is still $O(L^2)$, the Informer model further proposes a relative metric to estimate $M(q_i, K)$, defined as.

$$\bar{M}(q_i, K) = \max_j \frac{q_i k_j^T}{\sqrt{\sigma}} - \frac{1}{L_K} \sum_{j=1}^{N_K} \frac{q_i k_j^T}{\sqrt{\sigma}} \qquad (8)$$

Zhou et al. proved that under the long-tailed distribution, $\bar{M}(q_i, K)$ can be computed by simply randomly choosing $L_K \ln L_K$ pair of dot products. Eventually, the s query with maximum $\bar{M}$ is selected for the computation of self-attention, called probabilistic sparse self-attention mechanism:

$$A(Q, K, V) = softmax\left(\frac{\bar{Q}K^T}{\sqrt{d}}\right)V \qquad (9)$$

In Equation (9), the set $\bar{Q}$ contains s dominant query and $|\bar{Q}| = s$. If logL samples are chosen, the probabilistic sparse self-attention mechanism reduces the computational complexity of the traditional approach from $O(L^2)$ to $O(L \times \log L)$.

In addition, aiming to reduce the model computational load, the self-attention distillation layer is introduced for the streamlining of the output of the probabilistic sparse self-attention layer. This operation is implemented by employing a one-dimensional convolutional network and applying a function as the activation function; subsequently, the convolutional result is further compressed by a pooling layer with a step size of 2 allowing the original dimension of the probabilistic sparse self-attention output to be reduced by 50%. Finally, the output features from the multiple self-attention layers are integrated and concatenated into a single sequence, which is passed as an input to the decoder module.

### 3.2.3 Decoder

In the Informer model, its decoder consists of a combination of a fully connected layer, a fully attentive layer, and a multi-head probabilistic sparse self-attentive layer. The input to the structure is a fixed-length sequence of 0's of the relevant data of the past moments equal to the length of the predetermined output result. Informer's decoder realizes the same time series regression function as Transformer by means of masking the attention mechanism. Finally, the time series regression results are output through the fully connected layer of the decoder.

The objective function of the Informer model is MSE, defined as:

$$MSE = \frac{1}{L_y}\sum_{j=1}^{L_y}(l_{ij} - \hat{l}_{ij})^2 \qquad (10)$$

In Equation (10), $l_{ij}$ and $\hat{l}_{ij}$ denote the true and predicted values at the jth moment in the ith input sequence, respectively, and $L_y$ is the length of the sequence to be predicted. The training of Informer model parameters is achieved by minimizing the MSE.

### 3.3 Optimize wind speed period matching algorithm

The similar period method is widely used in the field of power load forecasting, in photovoltaic power generation, due to the photovoltaic power data is greatly affected by the weather, many scholars put forward the concept of "similar day", the photovoltaic power will be screened and classified according to the daily process, and the classification of different classification samples for classification modeling, to effectively improve the accuracy of the photovoltaic power prediction model by the concept of "similar day"[43]. Inspired by the concept of "similar day", when analyzing the wind power ramp event, the data of similar weather events in different moments in the same region can be used to play a reference role in the future wind power prediction and weather prediction[44], for example, in the field of photovoltaic (PV) power prediction, considering the cyclic nature of PV power generation, the weather conditions of the current period, such as temperature, rainfall and cloudiness, etc., may be similar to that of a certain day in the historical period, resulting in the similarity of PV loads. similarity, resulting in similarity of PV load similarity.

Inspired by the similar day concept of PV power generation, this paper applies the similar day concept to the field of wind power prediction, where a short-term sudden increase or decrease in wind power is often caused by a sudden change in wind speed in a sudden change meteorological environment, such as convective weather, transitory weather, etc., and the geographic location of the wind farm is fixed, and a sudden change that occurs at the current period also occurs in the historical wind power database, and the corresponding wind power curves will also have a correlation. Therefore, this study draws on the concept of similar days in the shape of daily load curves in photovoltaic power generation, extracts the convergence characteristics between the historical wind speed data, and proposes a similar period matching method based on wind speed correlation, and the specific algorithm is shown in Fig.11 Its realization method is divided into three steps:

Step 1: Calculate the climb coefficient of the pole period, different from several existing definitions of climb in the wind power field, this paper proposes a new definition of climb and introduces the climb coefficient.

Step 2: Find the similar matching sequences corresponding to each past wind speed segment in the historical wind speed using Fast Dynamic Time Warping algorithm (Fastdtw).

Step 3: Calculate the wind speed intensity difference and wind speed trend difference between the matched historical wind speed segments $w_a^h$ and the past wind speed segments $w_a^p$ and calculate the similarity coefficients between them.

Step 4: Summarize and align the two as features for subsequent prediction models.

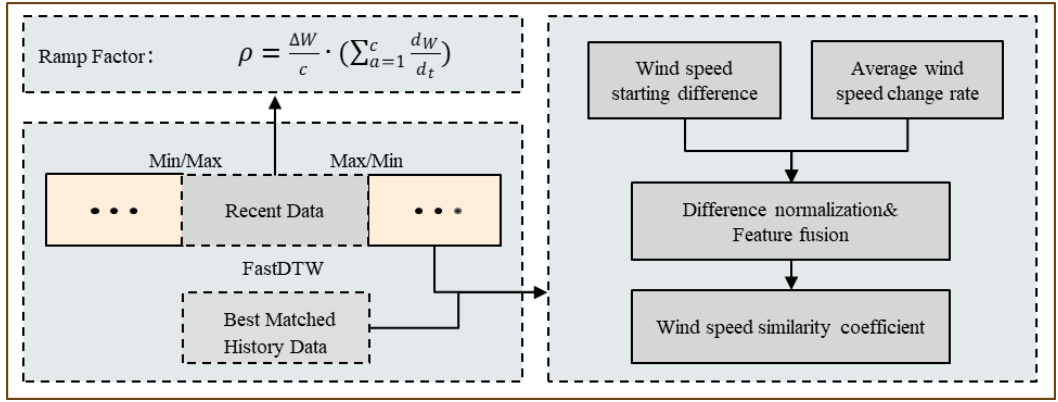

Figure 11. Algorithm for matching wind speed periods

**3.3.1 Ramp Factor (RF)**

For the definition of the ramp event in wind power, the industry does not have a clear definition of ramp, and currently it is mainly studied from the three aspects of the amount of ramp, duration, and direction of ramp, and four definitions are summarized in the literature[45], and there are two commonly used definitions at present:

Definition 1: For the selected period ΔT, whether the difference of wind power at the first and last end points is greater than the given threshold value $P_{val}$, if it is greater than the given threshold value, then it is considered that the climb occurs at the beginning of the time interval ΔT. That is, the formula is:

$$|P(t + \Delta t) - P(t)| > P_{val} \tag{11}$$

Definition 2: For the selected time interval ΔT, whether the difference between the minimum and maximum value of wind power is greater than the given threshold value $P_{val}$, if it is greater than the given threshold value, then it is considered that a ramp-up event occurs. That is, the formula is:

$$max(P[t, t + \Delta t]) - min(P[t, t + \Delta t]) > P_{val} \tag{12}$$

However, both existing definitions have certain limitations. Definition 1 lacks sensitivity in identifying ramp events characterized by rapid power fluctuations and continuous up-and-down ramp processes, while Definition 2 is less effective in detecting ramp events with extended durations. To address these shortcomings in the identification of ramp events, this paper introduces a novel concept: the definition of a ramp factor (RF method).

The preprocessed wind speed data is defined as two parts: historical wind speed data $w^h$ and past wind speed data $w^p$, where the past wind speed data refers to the processed wind speed data from November 6, 2023 to May 26, 2024 in Baijiashe area of Jiangxi Province, and the historical wind speed data refers to all the wind speed data of the Baijiashe area in the historical database in the 2 years before from the past wind speed data. The climb factor ρ is calculated for each wind speed segment by separating all the time segments according to the extreme points of the reconstructed wind speed series above, which is given by.

$$\rho = \frac{\Delta W}{c} \cdot \left( \sum_{a=1}^{c} \frac{d_W}{d_t} \right) \quad (13)$$

In Eq. (13), $\Delta W$ denotes the magnitude of wind speed change, a denotes the serial number of all time points in period c, c denotes the extreme point period, numerically the product of the number of time series points a and the fixed time step ΔT, ρ denotes the climb factor in period c, and $d_W/d_t$ denotes the climb factor for each point, numerically equal to the derivative of each time series point $a_i$. The specific principle is shown in Fig. 12 Shown.

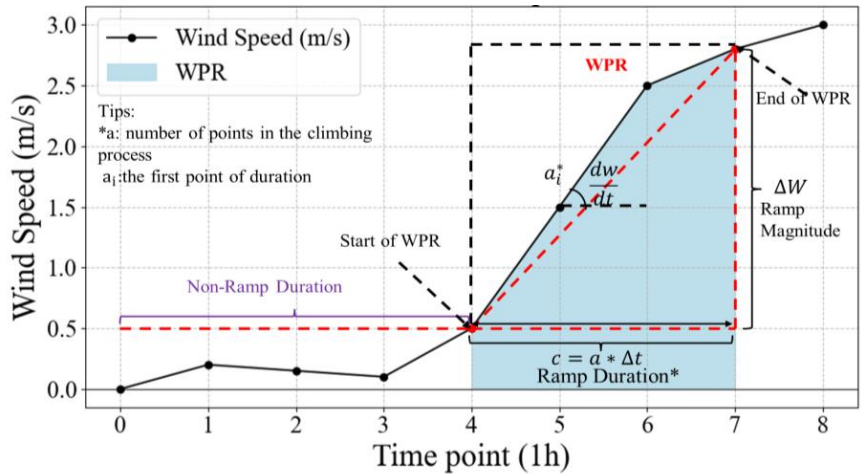

Figure 12. Definition of wind ramp factor

**3.3.3 Fastdtw algorithm**

To achieve wind power prediction based on the identification and matching of abrupt meteorological environments, it is crucial to extract features from historical wind speed data. This involves identifying historical wind power data segments that are similar to the predicted period and incorporating their features into the power prediction model. By referencing the features of historical wind speed segments during abrupt changes, the model can improve its accuracy in predicting wind power under similar meteorological conditions. In the domain of similar sequence matching, time series similarity distance analysis is a widely used method. Common approaches include Euclidean distance, dynamic time-warping (DTW) distance, and pattern distance. For analyzing the similarity of wind speed sequences, the DTW distance is utilized. As illustrated in Fig. 13, the DTW distance can be expressed as follows:

$$DTW(X_C, X_Q) = \min_p \sum_{k=1}^{K} d(p_k) \tag{14}$$

where X C and X Q represent time series. The d ($p_k$) represents the cost of bending. DTW uses dynamic programming to calculate the similarity between two time series, and the algorithm complexity is O($N^2$). When two time series are relatively long, the efficiency of DTW algorithm is slow and cannot meet the needs. Therefore, there are many algorithms to accelerate DTW: Fastdtw, Sparse-DTW, LB_ Keogh, LB_ Improved etc.

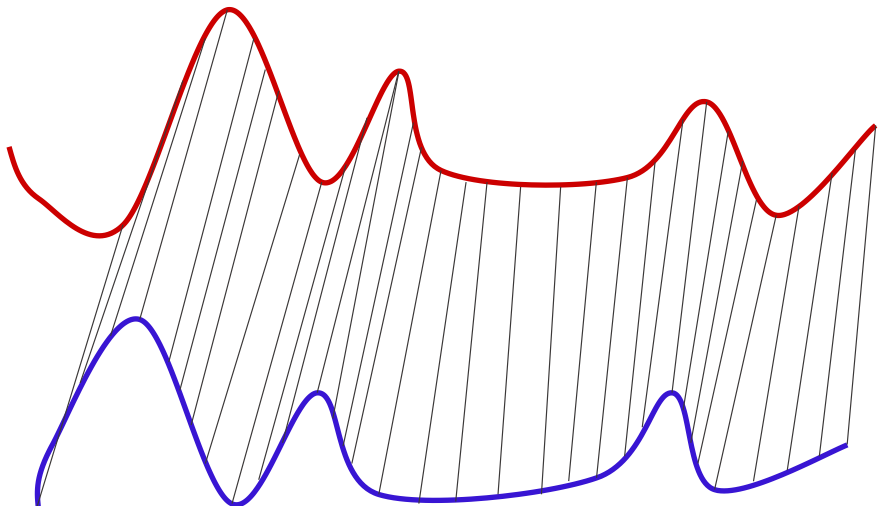

Figure 13. Dynamic Time Warping algorithm

To enhance the computational efficiency and ensure the applicability of the algorithm, this study employs the Fast-DTW algorithm. Fast-DTW integrates two techniques—constraint and data abstraction—to accelerate the calculation of DTW. The process is divided into three main steps:

(1) Coarse granulation. The original time series undergoes abstraction, where data abstraction is performed iteratively through stages such as 1/8 ->1/4 ->1/2->1/1. At each step, the coarse-grained data point is calculated as the average of the corresponding fine-grained data points.

(2) Projection. The DTW algorithm is applied to the coarse-grained time series.

(3) Fine-grained. After the coarse-grained path is determined, the algorithm further refines the squares passed by the reduced path, mapping them back to the fine-grained time series. Beyond this fine-graining process, the algorithm also extends the search area outward by M-granularity in transverse, vertical, and oblique directions in the fine-grained space. Here, M represents the radius parameter, typically set to 1 or 2[46]. The specific implementation of the Fast-DTW algorithm is illustrated in Fig.14. By employing this search-space reduction strategy, Fast-DTW achieves significantly lower computational complexity compared to traditional DTW algorithms, with a time complexity of O(N)[47].

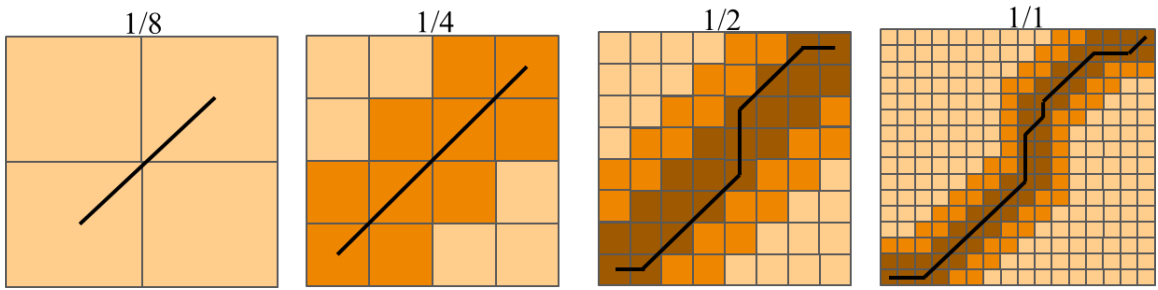

Figure 14. Fastdtw algorithm

### 3.3.4 Wind speed similarity coefficient

After the above improved Fastdtw algorithm matches each past wind speed segment to the most similar period data in the historical wind speed segment database, in order to better optimize the model features, a wind speed similarity coefficient $\Omega$ is used to show the wind speed correlation between the historical wind speed segment $w_a^h$ and the past wind speed segment $w_a^p$. In this paper, two parameters, wind speed intensity difference ($Wind_{STR}$) and wind speed trend difference ($Wind_{TRE}$), are introduced as follows:

$$Wind_{STR} = \frac{\sum_{a=1}^{c}|w_a^h - w_a^p|}{c} \tag{15}$$

$$Wind_{TRE} = \frac{1}{c-1}\left(\sum_{a=1}^{c-1}\frac{(w_{a+1}^h - w_a^h)}{\Delta t} - \sum_{a=1}^{c-1}\frac{(w_{a+1}^p - w_a^p)}{\Delta t}\right) \tag{16}$$

p denotes the historical wind speed data at the time a moment, $w_{a+1}^h$ denotes the past wind speed data at the ath+1th moment; the obtained wind speed intensity difference $Wind_{STR}$ and wind speed trend difference $Wind_{TRE}$ are normalized; where the normalized interval of the wind speed intensity difference $Wind_{STR}$ is set to (0,1), and that of the wind speed trend difference is set to (-1,1), then the wind speed similarity coefficient Ω is obtained by calculating the normalized wind speed intensity difference $Wind_{STR}$ and wind speed trend difference $Wind_{TRE}$, the specific formula is:

$$\Omega = \begin{cases} |Wind_{STR}^2 + Wind_{TRE}^2|, Wind_{TRE} > 0 \\ |Wind_{STR}^2 - Wind_{TRE}^2|, Wind_{TRE} < 0 \end{cases} \quad (17)$$

In Eq. (17), Ω denotes the wind speed similarity coefficient.

The climb coefficients of each extreme period and the matching completed past wind speed segments and historical wind speed segments and the wind speed similarity coefficients Ω of both are organized and merged. The similar period matching method based on wind speed correlation can predict the period as a reference to find convergent meteorological data from historical meteorological data, and avoid the problem of step size inconsistency in the DTW output results of traditional similarity matching method[48]. Meanwhile, the combined use of ramp coefficient and wind speed correlation is optimized based on the traditional ramp identification algorithm, and a new ramp identification method is proposed, which provides a new idea for solving the ramp time of wind power.

### 3.4 Wind power prediction method based on wind speed mutation event

In view of the same type of weather events at different times of the day have convergence characteristics can be effectively used for wind power prediction of wind speed mutation events, this paper proposes a wind power prediction method based on wind speed mutation event identification (hereinafter referred to as DMI), which consists of a combination of the wind speed mutation event identification module, wind speed similarity coefficient matching module and the Informer deep learning algorithm. The specific structure is shown in Figure 15, and the specific implementation steps are as follows:

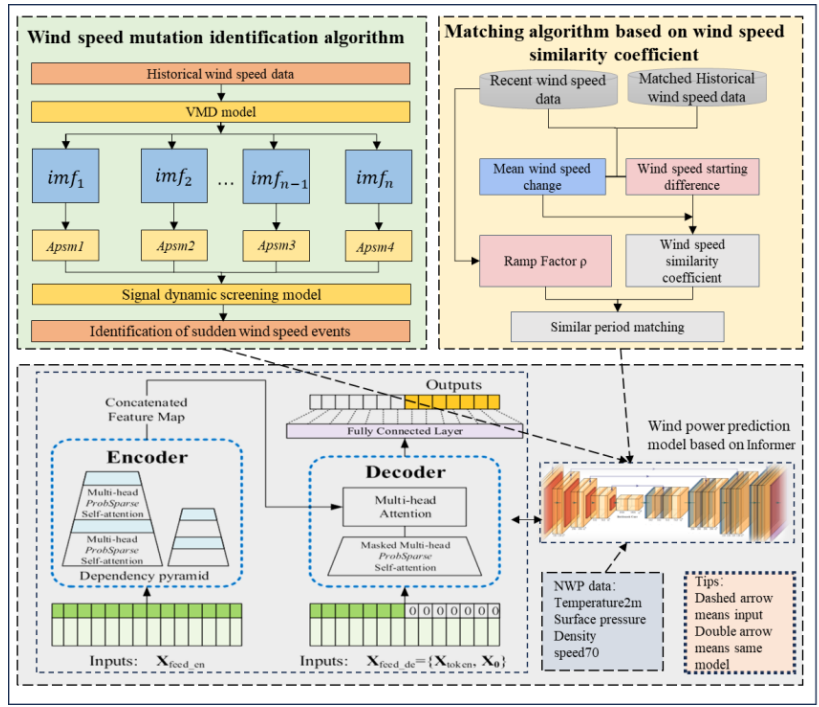

Figure 15. Structure of the power prediction method

Step 1: Optimized modal decomposition method for preprocessing wind speed data. The optimized VMD modal decomposition method after adding the pole adaptive selection model is used to decompose the raw data, and the decomposed component data are filtered and reconstructed according to the pole threshold (IC) proposed above to obtain the new reconstructed wind speed data. The optimized VMD algorithm proposed in this study is referred to as VMD-IC. Final, wind speed mutation events are identified and labeled.

Step 2: Similar matching of past wind speed segments to historical wind speed segments based on predicted wind speed data. Calculate the climb factor (RF) and wind speed similarity coefficient of the wind speed segment for the wind speed data segment identified and labeled in the previous step. Finally, an optimized Fastdtw algorithm incorporating the above two factors is used to achieve similar period matching.

Step 3: Construction of power prediction model. The matched meteorological data of similar time periods and their wind speed mutation event markers, the climb factor (RF) of each mutation period, numerical weather forecast data and past wind power data and other wind speed-power data are combined and constructed as multidimensional meteorological input data; finally, the constructed Informer deep-learning neural network is input to realize the wind power prediction.

### 3.5 Assessment methods

In addition to the errors stemming from the wind speed prediction process, the conversion of predicted wind speed into power predictions introduces another significant source of error. This section analyzes a real-world case of a wind power prediction system implemented at the target wind farm, utilizing power prediction evaluation indices relevant to practical engineering applications. The shortcomings of the current wind power prediction system in the wind power field are also discussed. This study enhances the commonly used error indicators for wind power prediction by incorporating the newly released national standards of the People's Republic of China. Additionally, new assessment indicators for wind farms are introduced. Specifically, the technical requirements for wind or photovoltaic power prediction systems adopted by the Jiangxi Province dispatch system serve as the performance indicators. These include the root-mean-square error (RMSE), mean absolute error (MAE), accuracy rate (AR), and power qualification rate ($PR_{Power}$).

The equations used for these metrics are provided as follows:

$$RMSE = \sqrt{\frac{\sum_{i=1}^{n}(P_{Mi} - P_{pi})^2}{n}} \tag{18}$$

$$MAE = \frac{1}{n}\sum_{i=1}^{n}|P_{Mi} - P_{pi}| \tag{19}$$

$$AC = \left(1 - \sqrt{\frac{1}{n}\sum_{i=1}^{n}\left(\frac{P_{Mi} - P_{pi}}{P_{Mi}}\right)^2}\right) \times 100\% \tag{20}$$

$$B_i = \begin{cases} 1, (1 - \frac{|P_{Mi} - P_{p,i}|}{C_i}) \geq 0.75 \\ 0, (1 - \frac{|P_{Mi} - P_{pi}|}{C_i}) < 0.75 \end{cases} \tag{21}$$

$$PR_{power} = \frac{1}{n}\sum_{k=1}^{n} B_i \times 100\% \tag{22}$$

In Equation (18)~(22), $P_{Mi}$ is the predicted value of wind power at time i; $P_{pi}$ is the measured value of wind power at time i; $C_i$ is the sum of the on-line capacity of the target wind farm at time i; $B_i$ is the qualified label of the power point at time i, with 1 being the qualified value of the point in the prediction, and vice versa; and n is the prediction time.

# 4. Results and discussion

To evaluate the effectiveness of the proposed method, this section conducts a detailed analysis across several factors. The processing results of the VMD-IC model, a VMD-based pole self-selection model, are discussed in detail in section 4.1. Section 4.2 presents the performance of the optimized Fast-DTW algorithm in matching predicted wind speed data with historical wind speed records. In Section 4.3, an ablation study is conducted across four datasets to validate the contribution of each module in the method. Finally, Section 4.4 describes multi-model comparative experiments, showcasing the superiority of the proposed method and its applicability in real-world engineering scenarios.

## 4.1 Self-selecting model for poles (VMD-IC model)

Considering the existence of "pseudo-poles" in the decomposed signals of VMD modal decomposition algorithm as shown in Fig. 17, i.e., the Euclidean distances between neighboring maxima and minima that meet the definition of poles are small and dense, this problem affects the selection of wind speed decomposition sub-data. In order to solve the above problem, the pole adaptive selection model is proposed and the wind speed decomposition data from November 6, 2023 to June 20, 2024 are selected as the experimental data, the VMD decomposition of the past wind speed data segments is carried out and then the poles are selected for each component, and the pole rate threshold $\mu_k = 1$ is set to find the poles of each component, and the specific pole rates are shown in the Table 1 and Fig.16 Shown, according to the pole rate threshold, the first three components are selected to merge and reconstruct, and the result of its partial time segment is shown in Fig. 17. The 98 mixed and closely spaced pole points of the original data segment in the figure are processed by the pole adaptive selection model to select 16 pole points for decomposition data selection. Finally, the first three components are reconstructed and analyzed to observe the extreme value points of the reconstructed data, and the reconstructed past wind speed data segments are shown in Fig. 18. Among them, the red circular points are the extreme value points, and the blue circular points are the very small value points, and the pseudo-polar point phenomenon is basically solved after observation.

In this study, the VMD model has been improved according to the above research methodology to form a customized VMD-IC model for optimization and denoising of wind speed data in wind power prediction

Table 1. Wind speed component pole rate

| VMD IMF | Number of extreme points | Rate of extreme points |
|---|---|---|
| 1 | 87 | 10% |
| 2 | 48 | 33% |
| 3 | 153 | 90% |
| 4 | 1432 | 196% |
| 5 | 1307 | 180% |
| 6 | 902 | 123% |
| 7 | 1718 | 235% |
| 8 | 1794 | 246% |

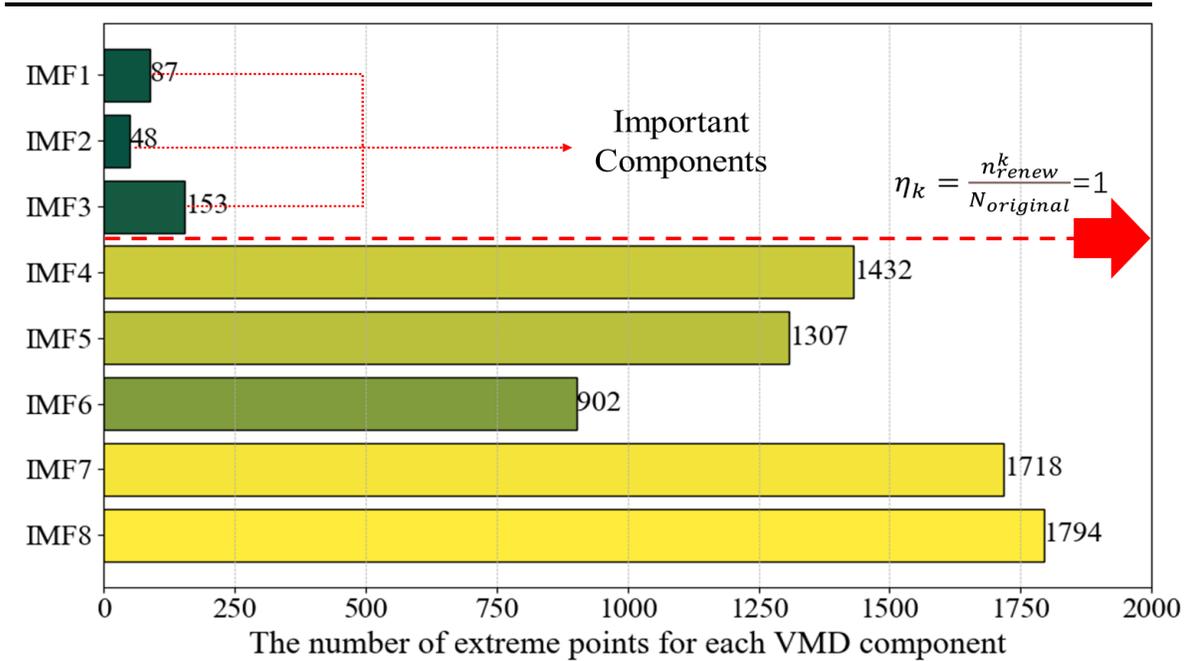

Figure 16. Number of extreme points

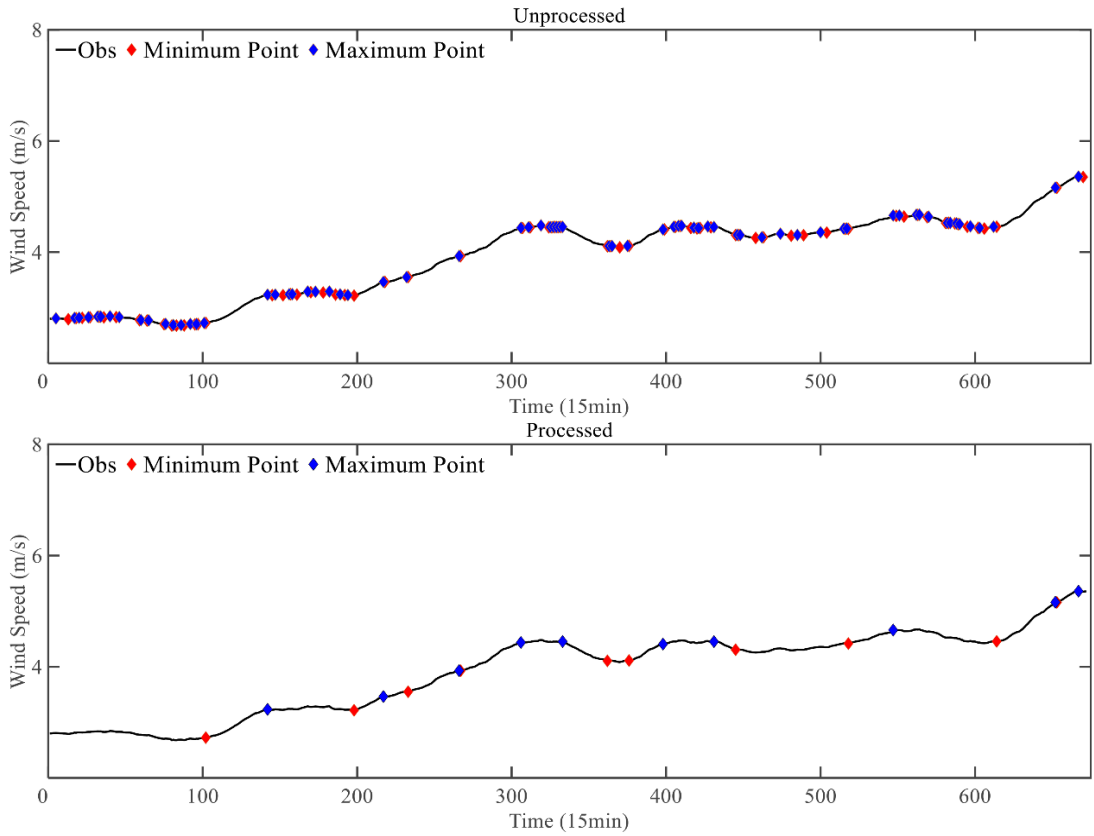

Figure 17. Pseudo pole removal

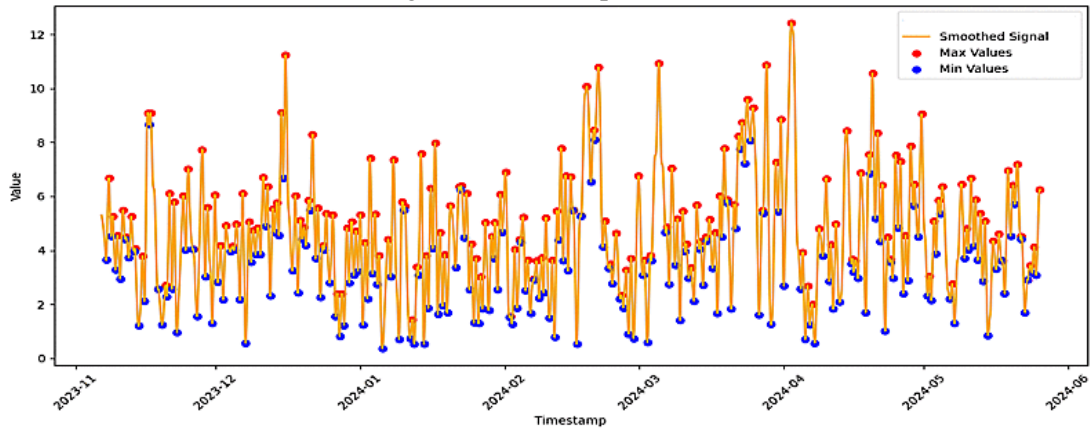

Figure 18. Reconstructed past wind speed sequence segments

## 4.2 Similar matching results

The wind power output and wind speed data from the Baijia She Wind Farm in Jiangxi, covering the period from May 2022 to May 2024, are set as past wind speed data, while the wind speed and power data from January 2018 to January 2022 for the same wind farm are designated as historical wind speed data. In the similarity matching process, the fastDTW algorithm is used to match similar past and historical wind speed data. To facilitate the analysis, the past wind speed data are divided into four datasets based on the four seasons: spring, summer, autumn, and winter. Each seasonal dataset is then compared with historical wind speed segments to identify the most similar wind speed sequences.

In Figure 19, subfigures (a-d) present selected ramp-up segments from the four seasonal datasets, while subfigures (e-h) show the matching results of ramp-down segments. Additionally, subfigures (i-l) display examples of matching results between past wind speed data and corresponding historical wind speed segments during wind speed oscillation periods. The experimental results presented in Figure 19 hold significant value for the subsequent phases of this research. By analyzing different wind speed scenarios, including ramp-up, ramp-down, and oscillation periods, the optimized fastDTW algorithm has demonstrated strong performance in identifying similar wind speed patterns across different time periods. This capability plays a crucial role in enhancing wind power forecasting, optimizing energy dispatch strategies, and improving the operational efficiency of wind farms.

The matching results clearly illustrate the correlation between past and historical wind speed data, enabling the forecasting model to effectively capture seasonal variations and transient wind characteristics. The segmentation approach based on seasonal patterns enhances the model's adaptability, allowing it to accommodate wind speed fluctuations unique to each season.These experimental findings lay a solid foundation for future research, offering empirical support for the application of data-driven techniques in wind power forecasting. The results validate the feasibility of the proposed methodology and provide practical guidance for further optimizing forecasting models and developing adaptive control strategies.

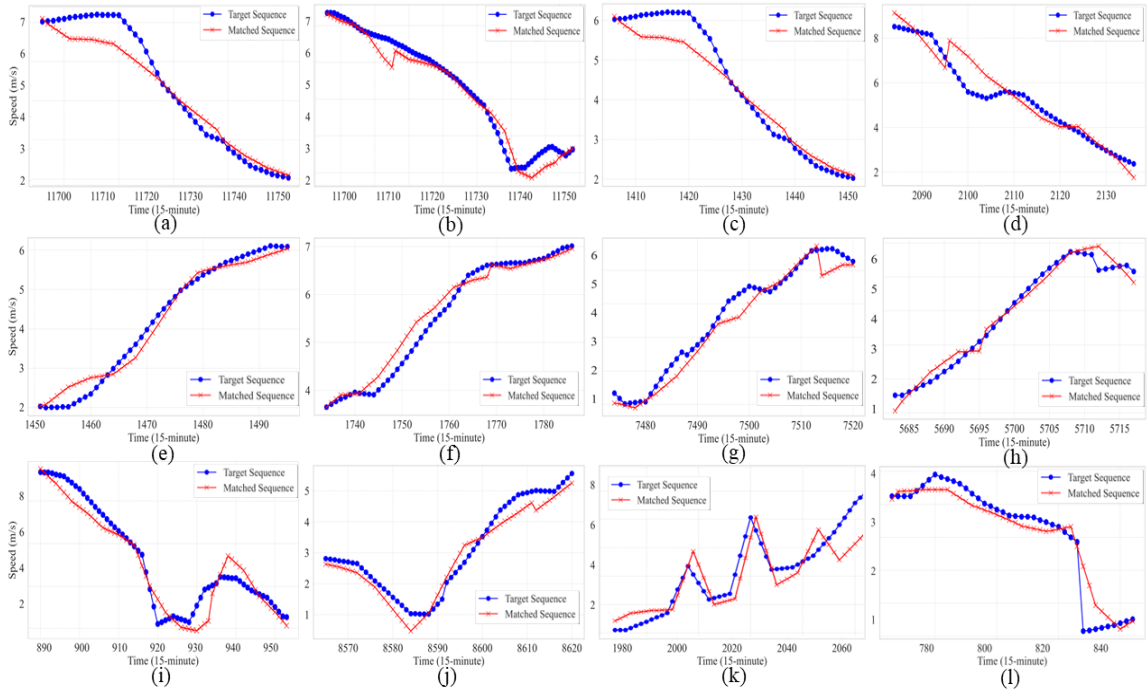

Figure 19. Ramp up, ramp down, oscillating range

## 4.3 Ablation study

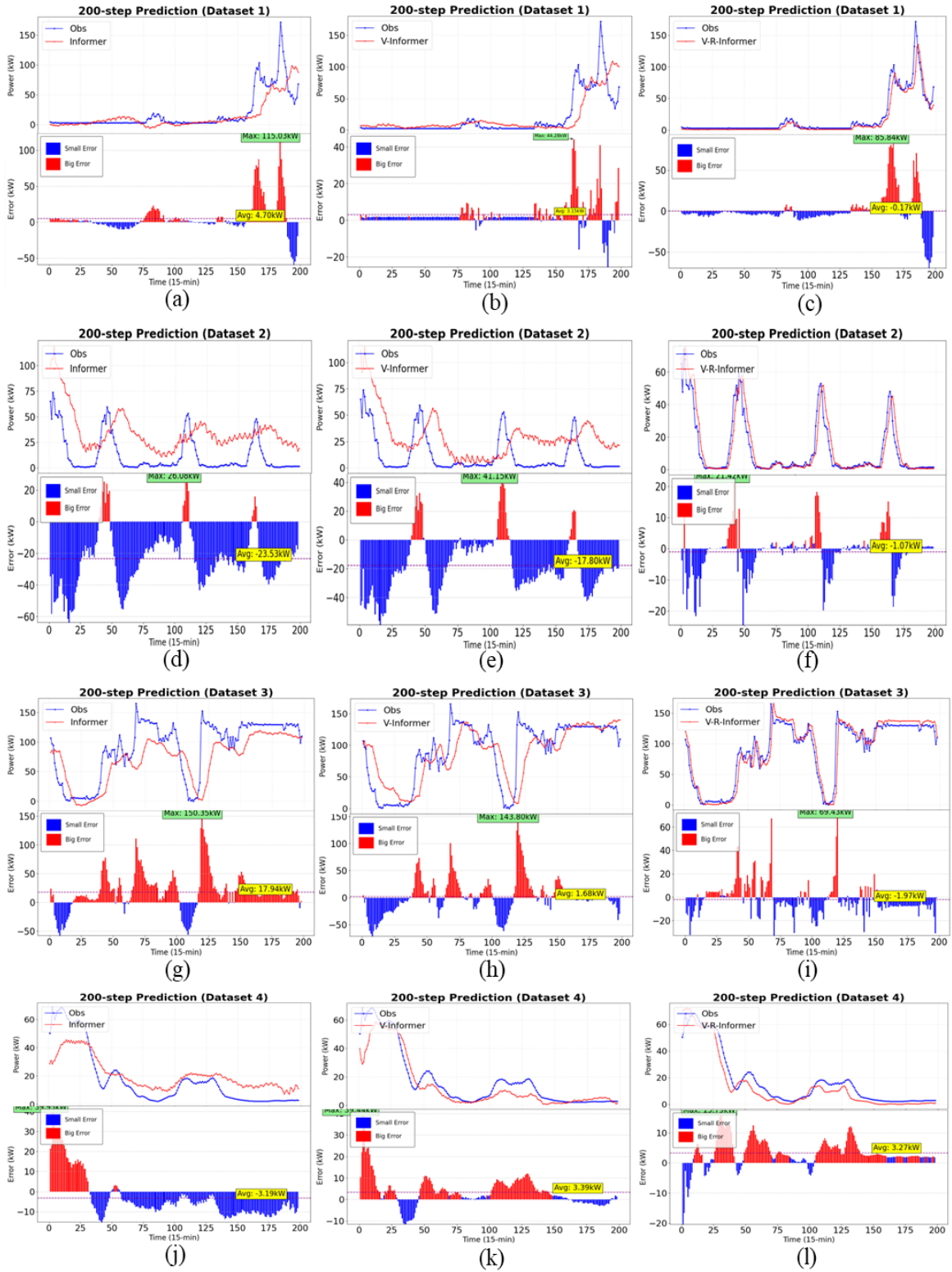

Figure 20. Ablation experiments with different data sets

In order to test the effectiveness of each module in this paper's power prediction modeling method, ablative experiments were conducted on the method, this test uses the hundred she wind farm data, one month is randomly selected from each season to conduct experiments, one month's dataset is trained to predict the power generation in the next two days, Fig 20 demonstrates the test of each model in four seasons, (the original Informer model in Fig 20. (a-c), (d-f), (g-i), and (j-l) respectively represent the test results of each model across the four seasons. Specifically:(a, d, g, i) correspond to the original Informer model;(b, e, h, k) correspond to the Informer model incorporating the pole adaptive selection mechanism and decomposed based on the VMD (IC) algorithm;(c, f, i, l) correspond to the Informer model further incorporating the similar model matching and climbing factor (RF).The Informer model decomposed by the VMD (IC) algorithm incorporating the pole adaptive selection mechanism proposed in the above study and the similar model matching Informer model incorporating the climbing factor (RF) are abbreviated as Informer, V-Informer, V-R- (Informer).

As shown in Fig 20, in the short-term wind power prediction for the next two days, the classical Informer model has the worst prediction effect, and the prediction will show obvious lag as well as missing magnitude when only the NWP data are passed in for power prediction; after processing the pole adaptive selection module based on VMD for each dataset, the lag of the predicted sequences is improved, and the identification of the climbing event plays a great progress, the power magnitude is more appropriate; furthermore, when the similar time period matching module with the optimized inclusion of the climbing factor is added, the accuracy is greatly improved, and the recognition rate of the climbing event is greatly improved.

The error assessment indices of each part of the model in the ablation experiments are shown in Table 2, which clearly shows that the original Informer-only model has lower indices in all four datasets, and with the addition of the VMD-IC module, the assessment indices are improved in the different training sets, and then with the addition of the creep factor (RF) module, compared to the VMD-IC-only Informer model, all the metrics in each dataset were also significantly optimized. In summary, the ablation experiments in this section validate the effectiveness of the various modules of the model used in this study in improving the accuracy of wind power prediction.

Table2 Error indicators for ablation experiments

| Model | Database1 | Database2 | Database3 | Database4 |
| --- | --- | --- | --- | --- |

| | | | | | |
|---|---|---|---|---|---|
| | V-R-Informer | 1.41 | 0.89 | 1.35 | 1.12 |
| $MAE$ | V-Informer | 1.88 | 1.06 | 1.71 | 1.37 |
| | Informer | 1.93 | 1.13 | 2.08 | 1.69 |
| | V-R-Informer | 40% | 20% | 31% | 37% |
| $r_{RMSE}$ | V-Informer | 53% | 25% | 39% | 46% |
| | Informer | 51% | 26% | 46% | 52% |
| | V-R-Informer | 31% | 16% | 25% | 30% |
| $r_{MAE}$ | V-Informer | 41% | 19% | 32% | 37% |
| | Informer | 42% | 20% | 39% | 45% |
| | V-R-Informer | 64.5% | 73.0% | 83.7% | 76.2% |
| $CC$ | V-Informer | 31.9% | 60.3% | 68.9% | 46.3% |
| | Informer | 56.6% | 52.1% | 67.1% | 69.8% |
| | V-R-Informer | 54.5% | 80.6% | 60.1% | 59.1% |
| $PR_{wind}$ | V-Informer | 44.8% | 73.3% | 56.6% | 53.5% |
| | Informer | 43.1% | 68.1% | 41.3% | 32.6% |

### 4.4 Comparison experiment of different models

To examine the performance of wind power prediction methods based on the identification of wind speed mutation events, this section selects several wind power prediction methods for performance comparison. In the current field of wind power prediction compared to ordinary machine learning algorithms, deep learning algorithms and hybrid model algorithms have more excellent performance in wind speed revision and wind power prediction. Therefore, in this section, SVM and XGBoost are chosen to represent traditional machine learning algorithms while deep algorithms such as LSTM, Transformer, and CNN-Transformer are used to represent hybrid models. The performance is compared with the VMD-IC-RF-I model proposed in this paper. The specific implementation method is to randomly 80% of the data from four datasets in spring, summer, autumn and winter for model prediction at the same time 200 time points are selected for prediction. The wind power prediction results are shown in Fig. 4.12, and the error metrics of the prediction results of each model are shown in Table 4-1. In addition, the proposed new method is noted as VMD-IC-RF-I as described above.

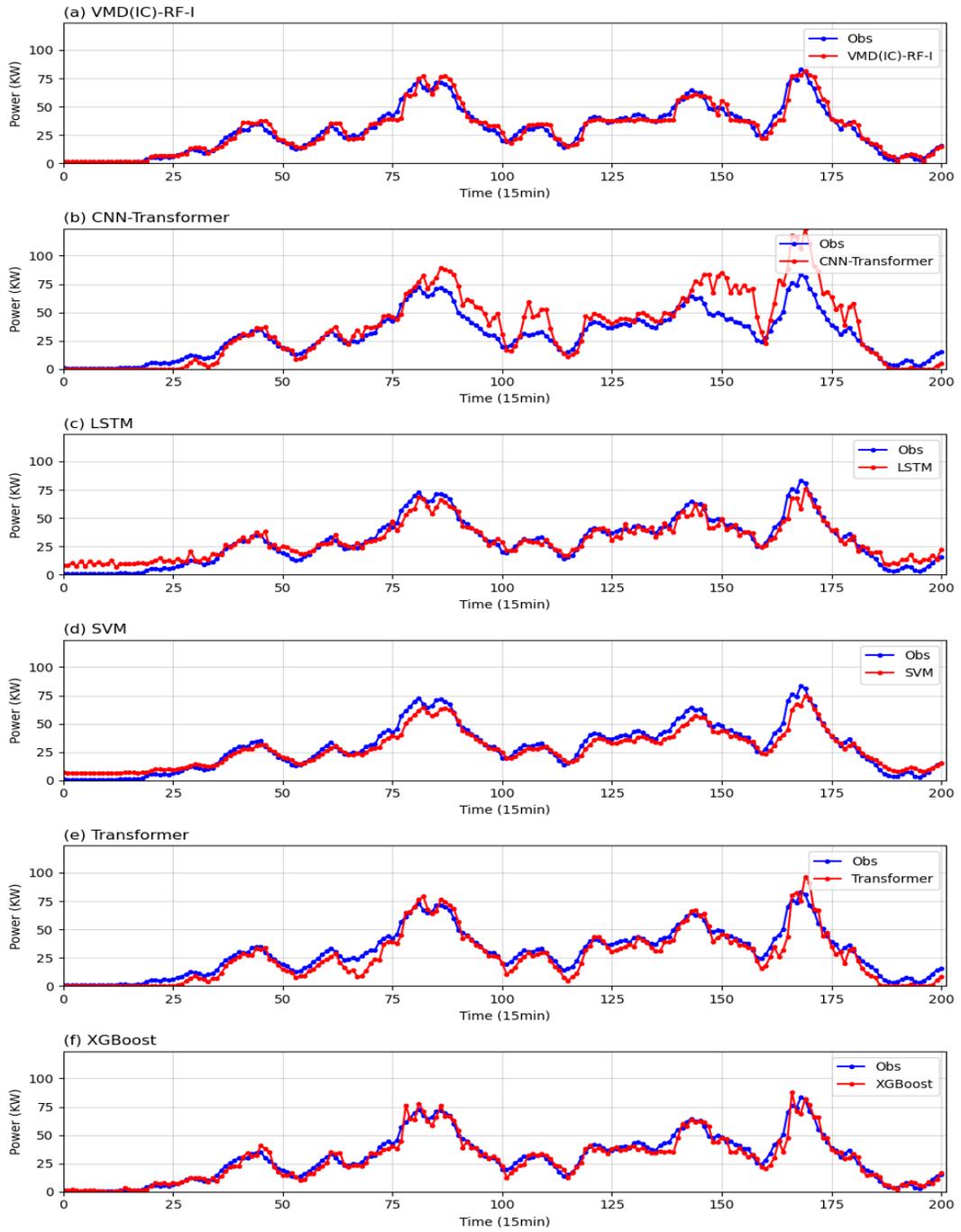

Figure 21. Prediction results of different models

Table 3 Evaluation indicators of different models

| Method | MAE | RMSE | AC | $PR_{power}$ |
|---|---|---|---|---|
| VMD(IC)-RF-I | 2.78 | 4.66 | 83.4% | 93.7% |
| SVM | 8.11 | 9.98 | 71.9% | 73.4% |
| XGBoost | 7.74 | 8.10 | 76.8% | 79.2% |
| LSTM | 2.96 | 5.23 | 77.6% | 82.6% |
| Transformer | 6.82 | 9.03 | 78.9% | 85.5% |
| CNN-Transformer | 5.74 | 7.83 | 80.4% | 89.9% |

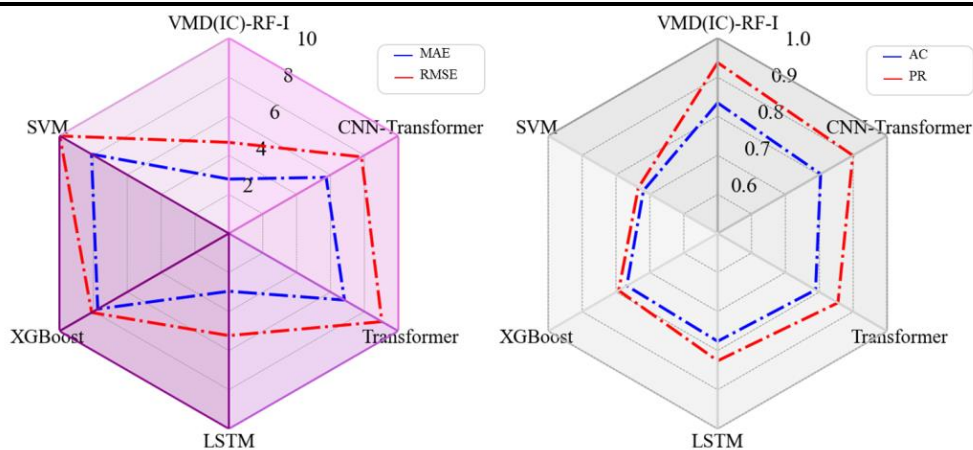

Figure 22. Evaluation Indicators Comparison

As shown in Figure 4.12, the power prediction curves of SVM neural network and XGBoost model cannot be ideally fitted with the actual power curve, which means that the traditional machine learning algorithms are also unable to cope with the task of wind power prediction for mountain wind farms. The wind power climbing problem caused by the sudden change of wind speed cannot be accurately predicted by traditional machine learning models. In addition, compared with the new deep learning algorithms led by Transformer and LSTM, the wind power prediction method based on the identification of wind speed mutation events has the best prediction performance, which can accurately characterize the stochastic, intermittent, and fluctuating nature of wind power data. Combined with Table 4-2 and Figure 4.13, the wind power prediction method based on the recognition of sudden meteorological environment has the smallest error index and the best accuracy and qualification rate, which means that the method can reduce the unqualified points of the wind power prediction and reduce the economic loss of wind power operators due to the insufficient performance of the power prediction algorithm. Similarly, compared with the hybrid model CNN-Transform, the model in this paper has significant advantages. In summary, from the perspective of power prediction, compared with many existing wind power prediction models, the wind power prediction method based on the identification of sudden meteorological events can effectively describe the future wind power curve by virtue of its excellent fitting ability; from the perspective of electric power practitioners, the method proposed in this paper can improve the accuracy rate of the wind farms to ensure the safety of the grid operation and at the same time, reduce the economic penalties incurred by power generators due to the low qualification rate.

From the perspective of electric power practitioners, the method proposed in this paper can improve the accuracy of wind farms to ensure the safe operation of the power grid and reduce the economic penalties incurred by power generation companies due to low qualification rates.

## 5. Conclusion

With the growing environmental challenges and increasing resource scarcity, green renewable energy sources with vast reserves, such as wind energy, have become a key focus in the global energy landscape. Wind energy, due to its accessibility and widespread distribution, has emerged as one of the most promising research areas among renewable energy sources. The large-scale adoption of wind energy can significantly reduce the reliance on traditional fossil fuels, contributing to a more sustainable energy structure. However, the stochastic, intermittent, and fluctuating nature of wind energy—amplified by factors such as terrain and seasonal variations—necessitates advanced wind power prediction methods to ensure its large-scale grid integration.

Considering one of the main reasons for the power prediction accuracy when the sudden change event of wind speed, this paper combines the Informer model with the similar day model to propose a climbing event correction model for short-term wind power prediction, which can accurately and effectively detect the climbing event based on the optimized similar period matching algorithm. The method eliminates the effect of 'pseudo-poles' caused by the fluctuation of the original wind speed series through the VMD-based pole adaptive selection model-VMD-IC model and realizes the selection of split-signal data from the VMD modal decomposition results for the identification of wind speed mutation events through the split-signal dynamic screening model.

In the wind speed mutation identification module, this study provides a definition that distinguishes from the existing wind speed climb identification, and proposes the climb factor (RF) formula as well as the wind speed similarity coefficient, which is combined with the concept of similar day in the field of photovoltaic power generation, and combined into an optimized similar time period matching algorithm to be applied in the identification of the climb time. Finally, the processed wind speed data is combined with multi-source data such as climb factor, wind speed correlation, and NWP weather forecast data to form the multi-modal data specific to this paper, which is combined with the new deep learning model Informer model to form the VMD-IC-RF-I model. In Section IV, ablation experiments are first conducted to verify the effectiveness of each model of the method, and finally, by comparing with the existing models such as LSTM, Transformer, SVM, etc., from the statistical data, compared with the other methods, the method of this paper has an MAE value of 2.78, an RMSE value of 4.66, and an AC and PR of 0.834 and 0.937, which are better than the rest of the models better than the rest of the models. It can be seen that the proposed method improves the performance of hill-climbing event identification as well as short-term wind power prediction, which can provide scientific guidance for economic and stable wind power prediction.

# References

[1] Yildiz C, Acikgoz H, Korkmaz D, Budak UJEC, Management. An improved residual-based convolutional neural network for very short-term wind power forecasting. 2021;228:113731.

[2] Meibom P, Weber C, Barth R, Brand HJIRPG. Operational costs induced by fluctuating wind power production in Germany and Scandinavia. 2009;3(1):75-83.

[3] Martinez A, Iglesias GJE. Global wind energy resources decline under climate change. 2024;288:129765.

[4] Khazaei S, Ehsan M, Soleymani S, Mohammadnezhad-Shourkaei HJE. A high-accuracy hybrid method for short-term wind power forecasting. 2022;238:122020.

[5] Gallego-Castillo C, Cuerva-Tejero A, Lopez-Garcia OJR, Reviews SE. A review on the recent history of wind power ramp forecasting. 2015;52:1148-57.

[6] Ferreira C, Gama J, Matias L, Botterud A, Wang J. A survey on wind power ramp forecasting. Argonne National Lab.(ANL), Argonne, IL (United States); 2011.

[7] Cui Y, Chen Z, He Y, Xiong X, Li FJE. An algorithm for forecasting day-ahead wind power via novel long short-term memory and wind power ramp events. 2023;263:125888.

[8] Aazami R, Heydari O, Tavoosi J, Shirkhani M, Mohammadzadeh A, Mosavi AJS. Optimal control of an energy-storage system in a microgrid for reducing wind-power fluctuations. 2022;14(10):6183.

[9] Ohba M, Kadokura S, Nohara DJRE. Impacts of synoptic circulation patterns on wind power ramp events in East Japan. 2016;96:591-602.

[10] Ahmed SD, Al-Ismail FS, Shafiullah M, Al-Sulaiman FA, El-Amin IMJIA. Grid integration challenges of wind energy: A review. 2020;8:10857-78.

[11] Yu G, Liu C, Tang B, Chen R, Lu L, Cui C, et al. Short term wind power prediction for regional wind farms based on spatial-temporal characteristic distribution. 2022;199:599-612.

[12] Celik AN, Kolhe MJAE. Generalized feed-forward based method for wind energy prediction. 2013;101:582-8.

[13] Santhosh M, Venkaiah C, Vinod Kumar DJER. Current advances and approaches in wind speed and wind power forecasting for improved renewable energy integration: A review. 2020;2(6):e12178.

[14] Ambach D, Schmid WJE. A new high-dimensional time series approach for wind speed, wind direction and air pressure forecasting. 2017;135:833-50.

[15] Liu H, Tian H-q, Li Y-fJEC, Management. Comparison of new hybrid FEEMD-MLP, FEEMD-ANFIS, Wavelet Packet-MLP and Wavelet Packet-ANFIS for wind speed predictions. 2015;89:1-11.

[16] Duan J, Zuo H, Bai Y, Duan J, Chang M, Chen BJE. Short-term wind speed forecasting using recurrent neural networks with error correction. 2021;217:119397.

[17] Ibrahim A, Mirjalili S, El-Said M, Ghoneim SS, Al-Harthi MM, Ibrahim TF, et al. Wind speed ensemble forecasting based on deep learning using adaptive dynamic optimization algorithm. 2021;9:125787-804.

[18] Yan X, Liu Y, Xu Y, Jia MJEC, Management. Multistep forecasting for diurnal wind speed based on hybrid deep learning model with improved singular spectrum decomposition. 2020;225:113456.

[19] Altan A, Karasu S, Zio EJASC. A new hybrid model for wind speed forecasting combining long short-term memory neural network, decomposition methods and grey wolf optimizer. 2021;100:106996.

[20] Zhao J, Guo Z, Guo Y, Lin W, Zhu WJE. A self-organizing forecast of day-ahead wind speed: Selective ensemble strategy based on numerical weather predictions. 2021;218:119509.
[21] Hwang BG, Lee S, Lee EJ, Kim JJ, Kim M, You D, et al. Journal of Wind Engineering and Industrial Aerodynamics. 2016;155:36-46.
[22] Zhao J, Guo Y, Xiao X, Wang J, Chi D, Guo ZJAe. Multi-step wind speed and power forecasts based on a WRF simulation and an optimized association method. 2017;197:183-202.
[23] Prósper MA, Otero-Casal C, Fernández FC, Miguez-Macho GJRe. Wind power forecasting for a real onshore wind farm on complex terrain using WRF high resolution simulations. 2019;135:674-86.
[24] Liu X, Lin Z, Feng ZJE. Short-term offshore wind speed forecast by seasonal ARIMA-A comparison against GRU and LSTM. 2021;227:120492.
[25] Cassola F, Burlando MJAe. Wind speed and wind energy forecast through Kalman filtering of Numerical Weather Prediction model output. 2012;99:154-66.
[26] Papaefthymiou G, Klockl BJItoec. MCMC for wind power simulation. 2008;23(1):234-40.
[27] Liu M-D, Ding L, Bai Y-LJEC, Management. Application of hybrid model based on empirical mode decomposition, novel recurrent neural networks and the ARIMA to wind speed prediction. 2021;233:113917.
[28] Groch M, Vermeulen J. Short-term ensemble nwp wind speed forecasts using mean-variance portfolio optimization and neural networks. Conference Short-term ensemble nwp wind speed forecasts using mean-variance portfolio optimization and neural networks. IEEE, p. 1-6.
[29] Hu J, Hu W, Cao D, Huang Y, Chen J, Li Y, et al. Bayesian averaging-enabled transfer learning method for probabilistic wind power forecasting of newly built wind farms. 2024;355:122185.
[30] Chen X, Zhao J, He M. Ensemble learning of numerical weather prediction for improved wind ramp forecasting. Conference Ensemble learning of numerical weather prediction for improved wind ramp forecasting. IEEE, p. 133-40.
[31] Chen Q, Chen Y, Bai XJE. Deterministic and Interval Wind Speed Prediction Method in Offshore Wind Farm Considering the Randomness of Wind. 2020;13(21):5595.
[32] Xiong X, Zou R, Sheng T, Zeng W, Ye XJE. An ultra-short-term wind speed correction method based on the fluctuation characteristics of wind speed. 2023;283:129012.
[33] Yang X, Xiao Y, Chen SJP-CSoEE. Wind speed and generated power forecasting in wind farm. 2005;25(11):1.
[34] Li J, Meng F, Zhang Z, Zhang YJER. Prediction of wind power ramp events via a self-attention based deep learning approach. 2024;12:1488-502.
[35] Li J, Song T, Liu B, Ma H, Chen J, Cheng YJIA. Forecasting of wind capacity ramp events using typical event clustering identification. 2020;8:176530-9.
[36] Dhiman HS, Deb DJIToEES. Machine intelligent and deep learning techniques for large training data in short‐term wind speed and ramp event forecasting. 2021;31(9):e12818.
[37] Al-Yahyai S, Charabi Y, Gastli AJR, Reviews SE. Review of the use of numerical weather prediction (NWP) models for wind energy assessment. 2010;14(9):3192-8.
[38] Wang Y, Zou R, Liu F, Zhang L, Liu QJAE. A review of wind speed and wind power forecasting with deep neural networks. 2021;304:117766.
[39] Abdoos AAJN. A new intelligent method based on combination of VMD and ELM for short term wind power forecasting. 2016;203:111-20.

[40] González-Cavieres L, Pérez-Won M, Tabilo-Munizaga G, Jara-Quijada E, Díaz-Álvarez R, Lemus-Mondaca RJTiFS, et al. Advances in vacuum microwave drying (VMD) systems for food products. 2021;116:626-38.
[41] He F, Zhou J, Feng Z-k, Liu G, Yang YJAe. A hybrid short-term load forecasting model based on variational mode decomposition and long short-term memory networks considering relevant factors with Bayesian optimization algorithm. 2019;237:103-16.
[42] Zhou H, Zhang S, Peng J, Zhang S, Li J, Xiong H, et al. Informer: Beyond efficient transformer for long sequence time-series forecasting. Conference Informer: Beyond efficient transformer for long sequence time-series forecasting, vol. 35. p. 11106-15.
[43] Shen Y, Zhang J, Liu J, Zhan P, Chen R, Chen Y. Short-term load forecasting of power system based on similar day method and PSO-DBN. Conference Short-term load forecasting of power system based on similar day method and PSO-DBN. IEEE, p. 1-6.
[44] Wang S, Zhang N, Wu L, Wang YJRE. Wind speed forecasting based on the hybrid ensemble empirical mode decomposition and GA-BP neural network method. 2016;94:629-36.
[45] D'Amico G, Petroni F, Vergine SJE. Ramp rate limitation of wind power: An overview. 2022;15(16):5850.
[46] Wu R, Keogh EJJIToK, Engineering D. FastDTW is approximate and generally slower than the algorithm it approximates. 2020;34(8):3779-85.
[47] Langfu C, Zhang Q, Yan S, Liman Y, Yixuan W, Junle W, et al. A method for satellite time series anomaly detection based on fast-DTW and improved-KNN. 2023;36(2):149-59.
[48] Choi W, Cho J, Lee S, Jung YJIa. Fast constrained dynamic time warping for similarity measure of time series data. 2020;8:222841-58.